\documentclass[conference]{IEEEtran}
\IEEEoverridecommandlockouts
\usepackage{cite}
\usepackage{amsmath,amssymb,amsfonts}
\usepackage{algorithmic}
\usepackage{graphicx}
\usepackage{textcomp}
\usepackage{xcolor}
\usepackage{hyperref}
\def\BibTeX{{\rm B\kern-.05em{\sc i\kern-.025em b}\kern-.08em
    T\kern-.1667em\lower.7ex\hbox{E}\kern-.125emX}}
\begin{document}

\title{An End-to-End Vehicle Trajcetory Prediction Framework\\

\thanks{Funded by The Hong Kong Polytechnic University College of Undergraduate Researchers \& Innovators (PolyU CURI)’s Undergraduate Research \& Innovation Scheme (URIS)}
}



\author{\IEEEauthorblockN{Fuad Hasan}
\IEEEauthorblockA{\textit{Department of Aeronautical \& Aviation Engineering} \\
\textit{The Hong Kong Polytechnic University}\\
Kowloon, Hong Kong SAR \\
frederick.hasan@connect.polyu.hk}
\and
\IEEEauthorblockN{Hailong Huang}
\IEEEauthorblockA{\textit{Department of Aeronautical\& Aviation Engineering} \\
\textit{The Hong Kong Polytechnic University}\\
Kowloon, Hong Kong SAR \\
hailong.huang@polyu.edu.hk}

}

\maketitle

\begin{abstract}
Anticipating the motion of neighboring vehicles is crucial for autonomous driving, especially on congested highways where even slight motion variations can result in catastrophic collisions. An accurate prediction of a future trajectory does not just rely on the previous trajectory, but also, more importantly, a simulation of the complex interactions between other vehicles nearby. Most state-of-the-art networks built to tackle the problem assume readily available past trajectory points, hence lacking a full end-to-end pipeline with direct video-to-output mechanism. In this article, we thus propose a novel end-to-end architecture that takes raw video inputs and outputs future trajectory predictions. It first extracts and tracks the 3D location of the nearby vehicles via multi-head attention-based regression networks as well as non-linear optimization. This provides the past trajectory points which then feeds into the trajectory prediction algorithm consisting of an attention-based LSTM encoder-decoder architecture, which allows it to model the complicated interdependence between the vehicles and make an accurate prediction of the future trajectory points of the surrounding vehicles. The proposed model is evaluated on the large-scale BLVD dataset, and has also been implemented on CARLA. The experimental results demonstrate that our approach outperforms various state-of-the-art models. 
\end{abstract}


\section{Introduction}

Recently, there has been a great deal of interest in autonomous driving systems, as it promises safer commute on the road. Among the numerous fundamental functions of autonomous driving, the prediction of the future trajectory of nearby vehicles has received special attention. When driving on a congested highway where cars frequently travel at high velocities, even a minor error can cause a chain reaction that results in a serious accident. In order to assess the danger of its own maneuvers, a self-driving vehicle must be able to predict the future paths of nearby vehicles.

In practice, however, it is exceedingly difficult to accurately predict the future trajectories of surrounding vehicles because it is dependent not only on their historical trajectories but also on the socio-temporal interdependence \cite{1} of the dense web of vehicular traffic surrounding the car. A model of the interaction between the participants must be used as one of the prediction parameters in order to make accurate predictions.

Initially, the majority of deep learning models concentrated on extracting the temporal correlation using RNN architectures such as LSTM and GRU \cite{20, 21}, which only considered the historical trajectory to make predictions for the future \cite{22}. Due to their encoder-decoder sequence-to-sequence architecture, RNNs were incapable of modeling social or spatial interactions. To address this issue, recent research \cite{44, 45} has explicitly concentrated on coupling RNNs with social feature extraction architectures to model vehicle interaction. Attention-based networks are a recent innovation due to their capacity to swiftly extract vital data from historical data. Two attention layers were stacked to extract both spatial and temporal attention in \cite{35} predicting the trajectory of pedestrians. Due to the problem of accumulating errors resulting from the autoregressive decoding procedure of Transformers \cite{9}, the multi-headed attention mechanism, originally proposed in the traditional Transformer \cite{7}, has not been extensively explored in the highway trajectory prediction problem despite these advancements.

In addition, the NGSIM dataset \cite{10} has been widely used to evaluate the validity of the vast majority of highway trajectory prediction studies. The limitation of implementing the NGSIM dataset is the fact that it contains readily available past trajectories. Most of the literature thus has exclusively focused on future trajectory prediction based on these past trajectories. To the knowledge of the authors, there has not been much work on a more robust, end-to-end pipeline that can both extract the 3D information from video sequences and also produce readily available past trajectories in order to extract the socio-temporal features and consequently predict future trajectories. An end-to-end pipeline makes it easier to deploy the solution on an actual autonomous vehicle as well, negating the dependence on 3D pose estimation architectures.

\begin{figure*}[t]
    \centering
    \includegraphics[width=\textwidth]{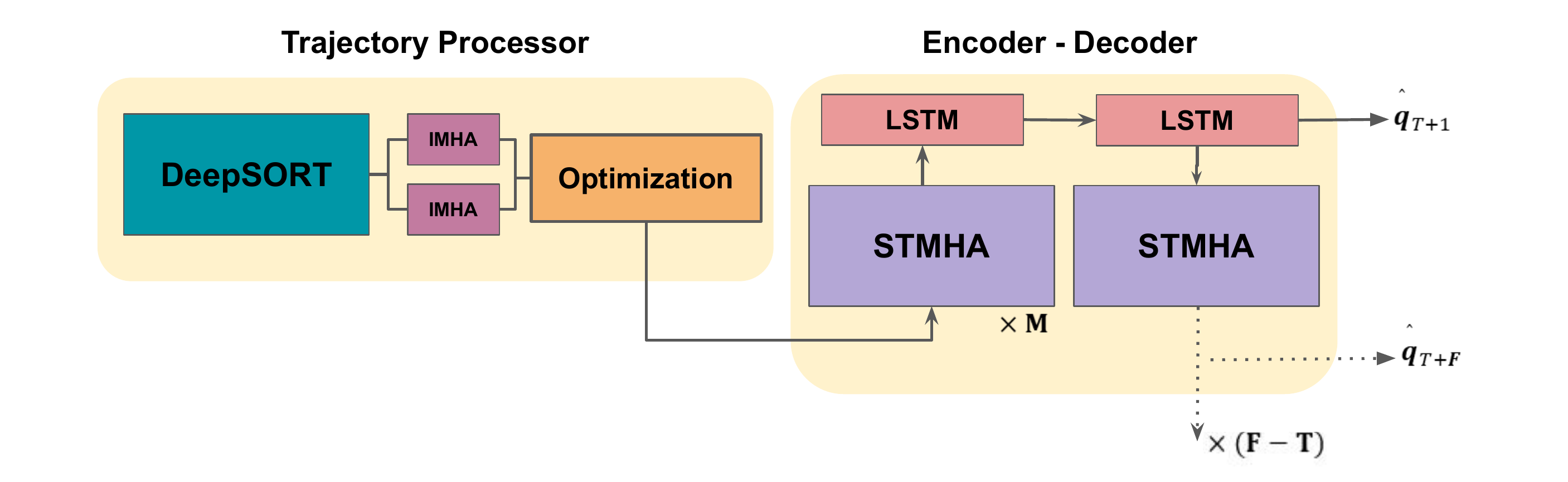}
    \caption{Overview of Proposed architecture} 
    \label{architecture}
\end{figure*}

Therefore, in this research, we propose an end-to-end vehicle trajectory prediction framework. We exploit the multi-head attention (MHA) mechanism of the Transformer to regress the 3D information from raw images and then fuse it with an LSTM sequence-to-sequence architecture which solves the decoder's autoregressive accumulative error problem.

The following is a summary of the contribution and academic novelty of our original work:

\begin{itemize}
    \item We propose a novel end-to-end architecture in order to produce predictions directly from video/image sequences.
    \item We utilize the MHA mechanism to extract 3D information from images, encode the trajectory in order to extract socio-temporal features and also extract future socio-temporal features.
    \item We evaluate our method on BLVD, a large-scale dataset of vehicle trajectories as well as video sequences. The experimental results demonstrate that our model strongly performs against state-of-the-art methodologies implemented on both the NGSIM and BLVD datasets, despite using video sequences as inputs.
\end{itemize}

\section{Methodology}

\subsection{Problem Formulation}

In this article, the problem of vehicle trajectory prediction has been formulated as a non-linear regression task where the inputs to the model are arbitrary video clip with frames $\left\{\mathcal{I}_1, \mathcal{I}_2, ... \mathcal{I}_n\right\}$ where $\mathcal{I}_i$ is a single image frame of the video clip. Based on the video, the task is to first extract the ego-centric coordinates of each vehicle $i$ in the frame as $p^i = (x^i, y^i)$. This can then be concatenated in order to produce the past trajectories of each vehicle represented as 
\begin{equation}
X=\left\{p_1, p_2, \ldots, p_T\right\}, 
\end{equation} 
where
\begin{equation} p_t = \left\{(x_t^0, y_t^0),(x_t^1, y_t^1),  \ldots, (x_t^N, y_t^N)
 \right\}. 
 \end{equation}
 $p_t^i = (x_t^i, y_t^i)$ are the coordinates $x$ and $y$ of vehicle $i$ at timestep $t$, where $t \in [1,T]$ and $i \in [1,N]$, $T$ stands for the total length of the observed trajectory and $N$ is the total number of observed neighbouring vehicles within 30 metres of the target vehicle, i.e. $\left|p_t^{target} - p_t^i \right| \le 30 $.

Based on the past trajectories, the objective of the proposed model is to learn a non-linear regression function that predicts the coordinates of all observed vehicles in the prediction horizon $F$, such that the predicted trajectories of the neighbouring vehicles are represented as $\hat{Y}$, as follows, which approximates the ground truth (GT) trajectory, $Y=\left\{{q}_{T+1}, {q}_{T+2}, \ldots, {q}_{T+F}\right\}$

\begin{equation}
\hat{Y}=\left\{\hat{q}_{T+1}, \hat{q}_{T+2}, \ldots, \hat{q}_{T+F}\right\},  
\end{equation} 
where
\begin{equation} \hat{q}_t = \left\{(\hat{x}_t^0, \hat{y}_t^0),(\hat{x}_t^1, \hat{y}_t^1),  \ldots, (\hat{x}_t^N, \hat{y}_t^N)
 \right\}.  
 \end{equation}

 The frame of reference of the dataset used in this paper is by default ego-centric as the trajectories were extracted from the onboard sensors. The longitudinal y-axis indicates the direction forward to the ego vehicle and the lateral x-axis direction indicates the axis perpendicular to it. The right-hand side is considered positive according to the dataset \cite{16}. As a result, the model is independent of the curvature of the road which conveniently allows it to be used in the highway as long as an object-detection and lane estimation algorithm is built on the target vehicle \cite{3}.

\subsection{Network Overview}

An overview of the model we propose is shown in Figure \ref{architecture}. It is divided into the following three modules: Trajectory Processor, Encoder and Decoder. We treat the surrounding vehicles as 3D bounding boxes. The Trajectory Processor first extracts the location of the centroid of these bounding boxes. We implemented  DeepSORT\cite{92f} in order to solve the Multi-Object Tracking (MOT) problem \cite{1p} first and get the 2D bounding boxes of surrounding vehicles. From the 2D bounding box patch provided by DeepSORT, we regress the dimension and orientations via two MHA layers to constraint and solve the optimization of the 3D real-world, ego-centric position of the 3D bounding boxes. This is then concatenated across frames to provide the historical tracks of the vehicles. The encoder module then encodes the socio-temporal attention data into the historical tracks via two distinct multi-head-attention layers and then also encodes temporal memory via the LSTM layer. The decoder then decodes the encoded information into future predictions. The hidden features are passed onto successive decoding steps via further MHA layers to improve further future predictions which also prevents the traditional transformer decoding accumulation error by enriching the hidden features.

\subsection{Trajectory Processor} 

 \begin{figure}[b]
    \centering
    \includegraphics[width=\linewidth]{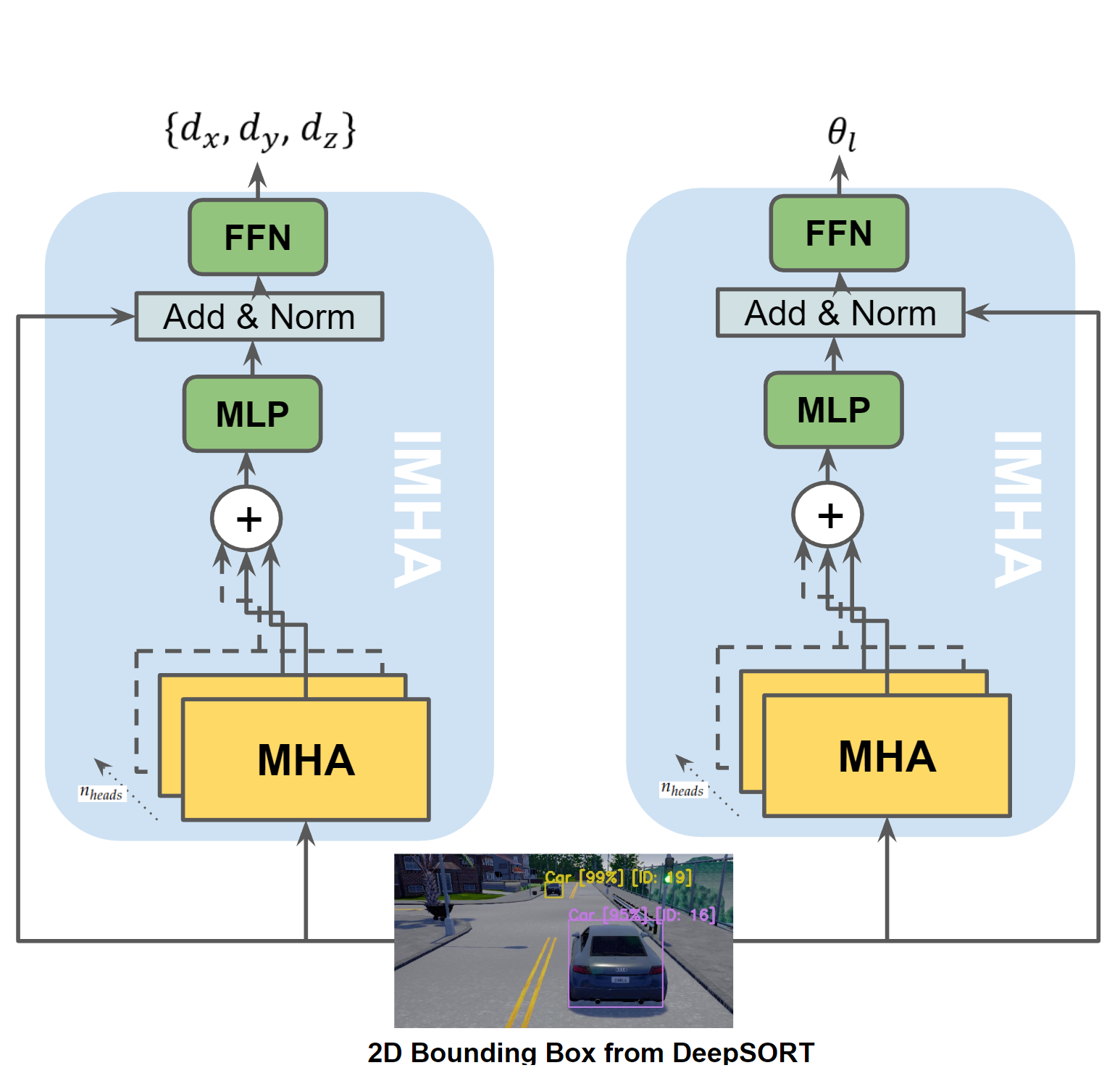}
    \caption{Skeleton of the IMHA layers}
\end{figure}

\subsubsection{Constraint Design}

 The 3D bounding box for surrounding vehicles is described by its translation/position $T=(t_x, t_y, t_z)$, dimensions $D=(d_x, d_y, d_z)$ and orientation $R=(\theta, \phi, \psi)$. The projection of a 3D point $X = (X, Y, Z, 1)^T$ into the image coordinate frame $x = (x, y, 1)^T$ with Camera Intrinsic $K$ is described as follows:
\begin{equation} \label{Constraint}
\mathbf{x}=K\left[\begin{array}{ll}
R & T
\end{array}\right] \mathbf{X}
\end{equation}

Taking the assumption that the center of the object coincides with the center of the 3D bounding box with known dimensions D, the vertex coordinates of the 3D bounding box is as follows: 

\begin{equation}
\mathbf{X_1} = (\frac{d_x}{2}, \frac{d_y}{2}, \frac{d_z}{2}),
\ldots \mathbf{X_8} = (-\frac{d_x}{2}, -\frac{d_y}{2}, -\frac{d_z}{2})
\end{equation}

In line with existing literature \cite{92f, 97f}, we further assume that the projection of the 3D bounding box fits tightly inside the 2D bounding box. This assumption requires that every side of the 2D box must contain at least one of the vertices of the projection of the 3D box. As an example, we can say that vertex $\mathbf{X_1} = (\frac{d_x}{2}, \frac{d_y}{2}, \frac{d_z}{2})$ lies on the left side of the 2D box with coordinate $x_{max}$. This correspondence of the point to the side of the 2D box can be represented in the following equation: 
\begin{equation} 
x_{\min }=\left(\left[\begin{array}{ll}
R & T
\end{array}\right]\left[\begin{array}{c}
d_x / 2 \\
-d_y / 2 \\
d_z / 2 \\
1
\end{array}\right]\right)_x
\end{equation}

Similarly, three other constraint equations can be derived for $x_{min}, y_{max}, y_{min}$, totaling to four constraint equations, so far. Further we make the assumption that the vehicles are upright, the relative roll of the vehicles are zero, the pitch of the vehicles are also zero and the driving scene is flat (highway), which reduces the possible configurations to only 64. Therefore, regressing the dimensions and orientations enables us to optimize the configurations in order to solve for translation $T$, from which past trajectories can be derived.

\subsubsection{Image Multi-Head Attention}

We implemented two Multi-Head Attention layers in the module Image Multi-Head Attention (IMHA) in order to regress the Dimension $d_x, d_y, d_z$ and the orientation $R=(\theta, \phi, \psi)$. As we assume roll $\phi$ and pitch $\psi$ to be zero, we only regress $\theta$ which is the yaw or azimuth angle of the vehicles. However, the visual appearance of the images correlates strongly with local orientation $\theta_l$ not the global orientation $\theta$, the two being related according to Equation \ref{or_eqn}. $\theta_{ray}$, which is the relative angle of the ray that intersects with a particular pixel, can be easily computed via the camera intrinsics. 
\begin{equation}\label{or_eqn}
    \theta = \theta_{ray} + \theta_l
\end{equation} The IMHA module thus directly outputs the dimension $d_x, d_y, d_z$ and local orientation $\theta_l$ (which consequently gives $R$) of the bounding boxes from the 2D image patch, which serves as inputs to Equation \ref{Constraint}. For example, let the vertical side of the 2D bounding box be $\mathbf{X^i}$, which is the $i^th$ corner of the 3D Box constrained in the 2D box, then the constraint equation is the following:

\begin{equation}
\left(K\left[\begin{array}{cc}
I & R \times \mathbf{X}^{\mathbf{k}} \\
0 & 1
\end{array}\right]\left[\begin{array}{c}
T_x \\
T_y \\
T_z \\
1
\end{array}\right]\right)_x=x_{\text {min}}
\end{equation}
where $I$ is a $3 \times 3$ identity matrix and $i$ corresponds to possible configurations of the 3D box corned projected on the 2D box as explained in \ref{Constraint}. The only unknowns remain to be the 3DoF relative position of the vehicles, $T=\left[T_x, T_y, T_z\right]$. As it is an over-determined set of linear equations with only one global minimum, it is solved by simple least square fitting optimization with a zero initial guess. The position coordinates are then concatenated to produce trajectory arrays. 

\subsection{Encoder-Decoder}
We have taken inspiration from \cite{mp} in order to design the Encoder-Decoder module in this section. To input the trajectory data into our proposed model, the coordinates are scaled first to a range of (-1, 1) to aid the model to reach faster convergence \cite{37}. This is then run through an embedding layer, a multi-layer perceptron (MLP) which, as shown in Eq. \ref{eq5}, maps raw input features to a higher dimension and outputs $X_{\lambda}$. 

\begin{equation} \label{eq5}
{\Lambda}_t^i=\operatorname{MLP}\left(p_t^i, W_{\lambda}\right)
\end{equation}  

The input is then fed into the Social-Multi-Head Attention Layer (SMHA). In this layer, the features of the input are first split into $n_{heads}$ transformer heads and fed into the Multi-Head Attention (MHA) layer. The MHA first computes the  query, key, and value matrices at timestep $t$, as shown in Eq. \ref{eq7}

\begin{equation} \label{eq7}
\left\{Q_t=\operatorname{MLP}\left(\left\{x_t^i\right\}_{i=1}^N, w_q\right), \quad K_t, \quad V_t\right\}
\end{equation}

The inter-vehicular interaction is represented by an undirected graph $G_t = \left\{ V_t, E_t\right\}$ where the nodes $V_t$ represent the observed vehicles $V_t = \left\{ v \mid i = 1, 2, ..., N \right\}$ and the edges represent the interaction as binary values between vehicles $i$ and $j$, $E_t[i][j] = 1$ if $p_t^i - p_t^j \le d_{near} $, otherwise the value is zero. The interaction graph is used to mask social attention. 

The masked multi-head attention of head $h$ at timestep $t$ is then computed as
\begin{equation}
\operatorname{A}_h\left(Q_t, K_t, V_t\right)=\frac{\operatorname{Softmax}\left(\left[\left(q_t^i\right)^Tk_t^j\right]_{\Gamma}\right)}{\sqrt{d_k}}\left[v_t^i\right]_{\Gamma},
\end{equation} 
\begin{equation}
\operatorname{MHA}\left(Q_t, K_t, V_t\right)=\operatorname{Concat}\left(A_1, ...,A_{n_{heads}}\right).
\end{equation} where $\Gamma =\left\{j \mid G_t[i, j]=1, j \in[1, N]\right\}$ is the attention mask. The output from the Multi-Head Attention layer of size $O_{MHA}$ is then fed through an intra-layer MLP which helps boost training speed. This is then followed by an Add \& Norm layer. The Add layer is simply a residual connection of the input $X_{\lambda}$ which adds the current output with the input. This connection greatly helps models such as transformers reach convergence faster by resolving the vanishing gradient problem so it is also used extensively in our model \cite{39}. The output of the Social Multi-Head Attention Layer is thus computed as follows:

\begin{equation}
    \operatorname{Norm}(\operatorname{MLP}(O_{MHA}) + X_{\lambda}).
\end{equation} 

Positional Encoding, via a simple MLP is then added to the output which is then fed through a Temporal Multi-Head Attention (TMHA) layer. The TMHA layer has a similar architecture as SMHA along with a time mask that prevents the current steps from accessing features from the relative future. For example, if the current timestep is $t$ then the features from timestep $t+1$ to $T$ will be masked to zero. This prevents the model from overfitting and making exclusive correlations on the training data. The output is also fed through another multi-layer perceptron network (FFN). FFN is made up of two feed-forward networks that successively map the feature vectors to a higher dimension and then back to the original dimension. This is done to add more model parameters so that the temporal attention vectors can be fed through further layers such as the LSTM. Thus the output from the TMHA layer has both social and temporal correlation encoded into it. We refer to the SMHA, TMHA, and FFN layers compounded together as Socio-Temporal Multi-Head Attention (STMHA) layer, as illustrated in Figure \ref{architecture}. We then stack $M$ number of STMHA layers successively to extract more complex socio-temporal dependence from the past trajectory information. 

We propose a traditional LSTM encoder after the first STMHA layer stack to extract the hidden memory information from the output tensor. This is done to facilitate the hidden tensor passing to the LSTM decoder module. We propose this sequence-to-sequence architecture as a replacement for the traditional transformer architecture with an MHA-based transformer decoder. As mentioned before, due to the autoregressive nature of the transformer decoder, it carries the problem of accumulated errors. To decode the hidden-state tensor and at the same time extract the rich socio-temporal dependence information encoded into the hidden tensor from the decoded information, we propose an STMHA-LSTM decoder. The hidden tensor at timestep $T+1$ is then fed into an STMHA layer to extract and update the socio-temporal dependence at the next decoding step. Such is repeated for every following decoding step. This is done to further improve the accuracy of successive decoding steps, as well as decode the future trajectories from the hidden information of the transformer encoder with lower auto-regressive accumulated errors.

\section{Experiments}

\subsection{Implementation Details}
\subsubsection{Dataset}The proposed model was trained, validated, and tested on the BLVD dataset. It consists of a total of 654 high-resolution videos which were extracted from Changshu city, Jiangsu province. Each frame consists of the ID, 3D coordinates, vehicle direction, and interactive behavior information of all observed vehicles by the ego. These frame annotations were used to train, test and evaluate our coordinate estimation algorithm and produce trajectories which were then run through the encoder-decoder to make future predictions. The prediction was then evaluated against the trajectory ground truth of the BLVD dataset.  In line with the existing literature \cite{1, 3, 35} we have adopted the root mean squared error (RMSE) between the prediction and the ground truth as our evaluation metric. RMSE at prediction time $t^{\prime}, t \in [T+1, ..., T+F]$ can be calculated as illustrated in Eq. \ref{rmse}. 

\begin{equation}\label{rmse}
\operatorname{RMSE}_t=\sqrt{\frac{1}{L} \sum_{l=1}^L\left(\hat{Y}_t^{l} -Y_t^{l}\right)^2},
\end{equation} where $\hat{Y}$ is the predicted positions and $Y$ is the Ground Truth Position of the \textit{l}-th testing sample at timestep $\textit{t}^{\prime}$ and L is the total length of the test set. 

\subsubsection{Hyperparameter Settings}
For the observable region, we set it to be a 30m radial region. We set the threshold $d_{near}$ to be 15m. We set the number of STMHA layers $M = 2$ and $n_{heads}$ of all the MHA modules to be 4. For the trajectory, we have used the past timesteps $T$ to be 3 s and future timestamps F to be 5 s. For model training, we have used the Adam \cite{40} optimizer with $\eta$ = 0.001, $\beta$ = 0.999. The learning rate used is 0.0001 and the batch size of 32. The teacher-forcing ratio used is 0.5.

\subsection{Ablative Tests}
To defend the effectiveness of our architecture, we perform three ablative experiments with the following three variants of our architecture
\begin{enumerate}
\item	\textbf{TP: } Excludes the Trajectory Processor and uses trajectories extracted from the dataset to test and validate.
\item	\textbf{EST: } Excludes the STMHA in the Encoder.
\item	\textbf{DST: } Excludes the STMHA in the Decoder.
\end{enumerate}
Following Table \ref{tab1} are the results for the above ablative setup.

\begin{table}[h]
\caption{Ablative Analysis}
\begin{center}
\begin{tabular}{c c c c c c c}
\hline
\textbf{Excluded}&\multicolumn{5}{c}{\textbf{RMSE}} & \textbf{Average} \\

\textbf{Module}& \textbf{1s}& \textbf{2s}& \textbf{3s}& \textbf{4s}& \textbf{5s}& \textbf{Improvement} \\
\hline
TP	& 0.48    & 1.01  & 1.60  & 2.31  & 3.36    & -29.6\% \\
EST	& 0.67    & 1.48  & 2.10  & 3.03  & 4.70  &  34.8\% \\
DST	& 0.64    & 1.61  & 2.36  & 3.35  & 5.66 & 67.6\% \\ \hline
Control& 0.57    & 1.32  & 1.86  & 2.66  & 3.82  & \\ \hline
\end{tabular}
\label{tab1}
\end{center}
\end{table}

At first, the results for the improvements in the EST and DST imply the effectiveness of the STMHA layer. The Encoder STMHA layer proved to be crucial in encoding socio-temporal interaction, as evident by the increase in error through all timesteps resulting from its exclusion (EST). The Decoder STMHA layer seems to play a vital role in reducing errors as well especially in future timesteps as evident by the exponential rise in error due to its exclusion in DST. Secondly, the exclusion of TP and directly importing the annotated trajectories of the dataset seems to positively impact the model, however, the error between the control (with TP) and the variant without TP does not seem that significant, with an average improvement of only -29.6\%. This suggests that although TP cannot compare to manually constructed annotations as in the BLVD dataset, it does not produce 3D coordinate estimates that are too far off. 

\begin{figure}[t]
    \centering
    \includegraphics[width=0.8\linewidth]{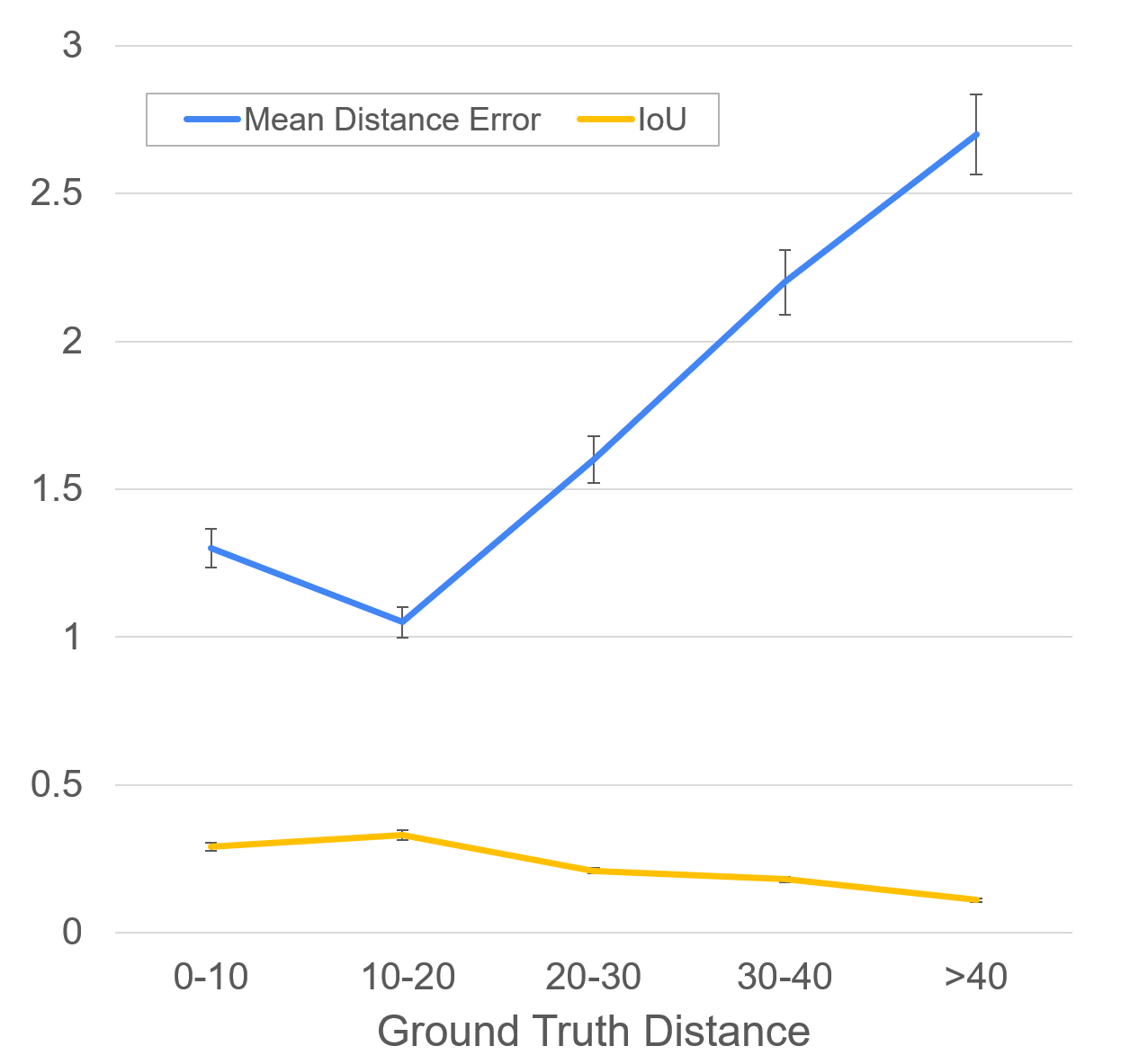} 
    \caption{MDE \& IoU of Target vs Distance from Ego}\label{iou}
\end{figure}

We also studied the mean distance error (MDE) (in meters) of the position estimation and the Intersection over Union (IoU) of the bounding box estimated vs the distance of the target from the ego. Figure \ref{iou} summarizes the findings.

\subsection{Comparison with Other Models}
Table \ref{tab2} summarizes our method's performance with the existing literature. It demonstrates our model's strength, especially in enhancing the prediction accuracy in future timesteps. 

\begin{table}[h]
\caption{Comparative Analysis}
\begin{center}
\begin{tabular}{c c c c c c}
\hline
\textbf{Model}&\multicolumn{5}{c}{\textbf{RMSE}}  \\

\textbf{}& \textbf{1s}& \textbf{2s}& \textbf{3s}& \textbf{4s}& \textbf{5s} \\
\hline
CV	& 0.73 & 1.78 & 3.13 & 4.78 & 6.68 \\
V-LSTM	& 0.68 & 1.65 & 2.91 & 4.46 & 6.27 \\
S-LSTM	& 0.65 & 1.31 & 2.16 & 3.25 & 4.55 \\
CS-LSTM  & 0.61 & 1.27 & 2.09 & 3.10 & 4.37 \\
DSCAN   & 0.58 &  1.26 & 2.03 & 2.98 & 4.13 \\
SGAN    & 0.57 & 1.32 & 2.22 & 3.26 & 4.40 \\
HMNet   & 0.50 & 1.13 & 1.89 & 2.85 & 4.04 \\ \hline
Ours & 0.57    & 1.32  & 1.86  & 2.66  & 3.82  \\ \hline
\end{tabular}
\label{tab2}
\end{center}
\end{table}

 \begin{figure}[t]
    \centering
    \includegraphics[width=\linewidth]{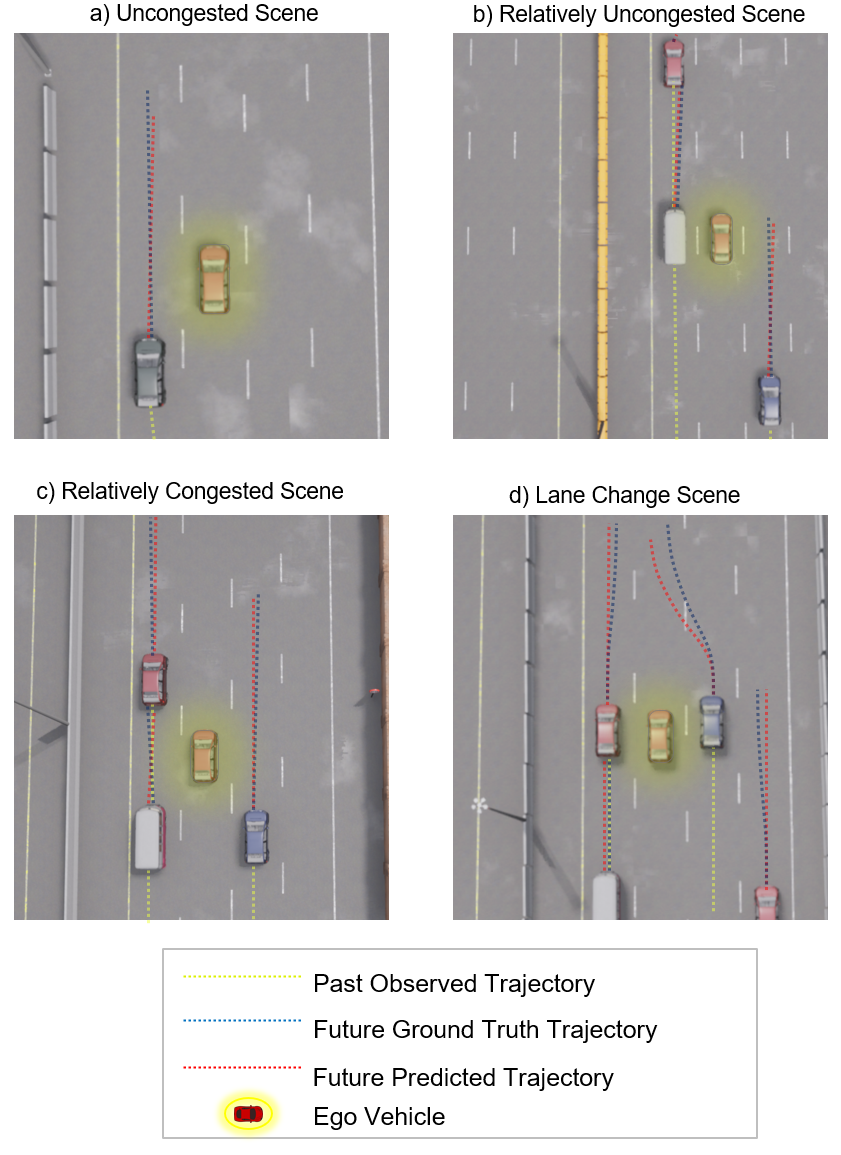} 
    \caption{Visualization of Real-time Trajectory Prediction on CARLA}\label{pred-vis}
\end{figure}
\subsection{Visualization}
The trained model was also implemented on CARLA's \cite{110f} Town10 in order to qualitatively validate the strengths and weaknesses of the model through visualized scenarios. Figure \ref{pred-vis} demonstrates four distinct scenes each with different levels of surrounding traffic, with the trajectories marked in different colors. The performance of the model seems similar for the first three scenes, with very low absolute path error. Our model was designed to deal with congested traffic by extracting socio-temporal dependencies and making a prediction, and this is further evident by these three scenarios. However, as can be seen in Figure \ref{pred-vis}(d), the absolute path increasingly deviates from GT in lane change scenarios, even though the model predicts an accurate direction of the lane change. We believe this is due to the highway dataset of BLVD not having enough lane change cases, which causes a sort of underfitting of the model's prediction in those scenarios.

\begin{figure}[t]
    \centering
    \includegraphics[width=\linewidth]{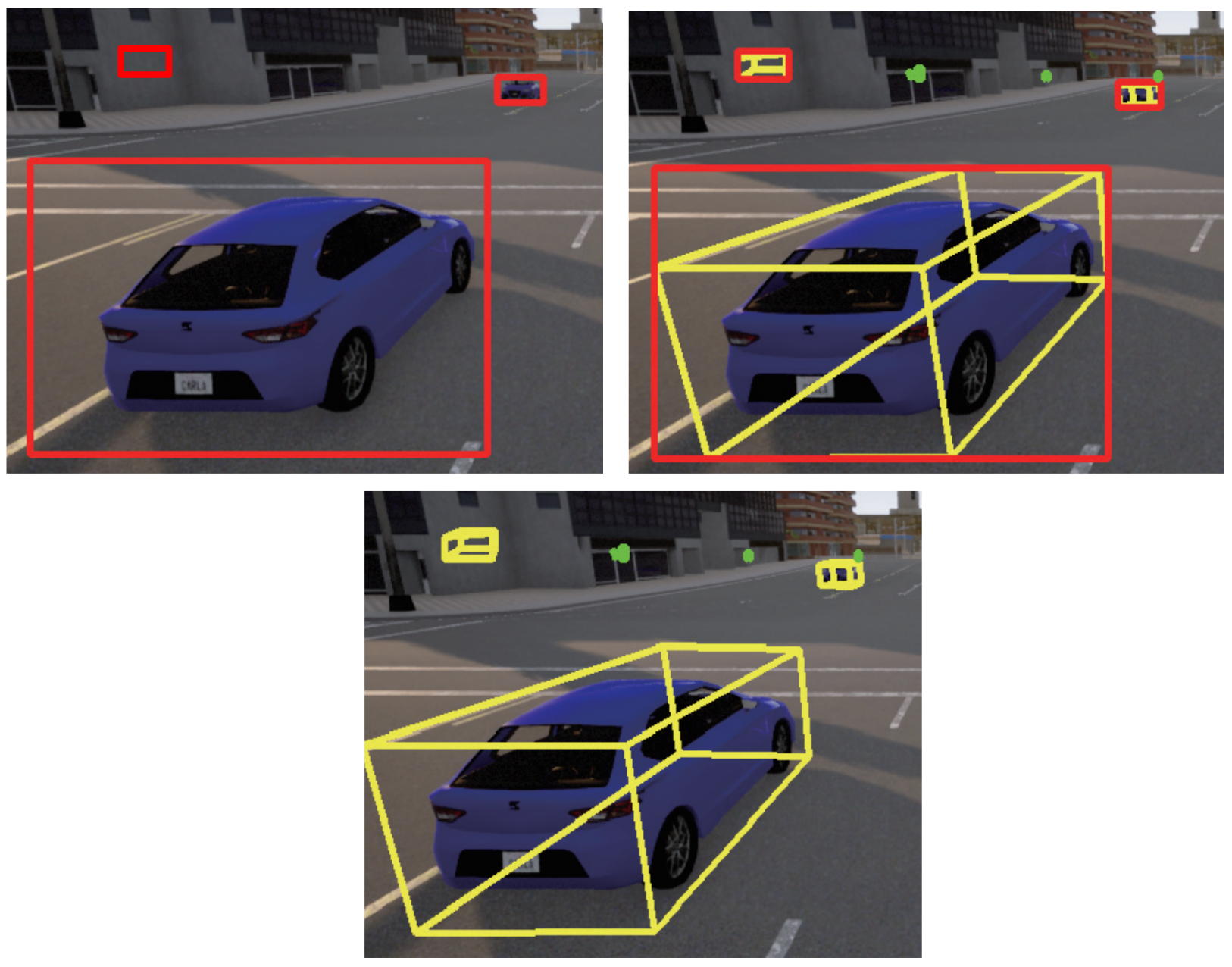} 
    \caption{Visualization of Bounding Box Reconstruction on CARLA}\label{box-vis}
\end{figure}

Visualization tests were also carried out to check the performance of the 3D bounding box prediction, as illustrated in Figure \ref{box-vis}. This demonstrates our technique of converting from 2D to 3D bounding box via the Trajectory Processor module. A wrong 2D prediction resulting in a faulty bounding box can be observed in this scenario (on the building to the left). Our method relies strongly on the 2D box prediction in order to estimate its 3D counterpart, which can be regarded as one of its limitations. So, if the 2D box is inaccurate, this would also propagate the errors to the 3D box. 

\section{Conclusion and Future Work}
This paper has introduced an end-to-end trajectory prediction framework, enabling trajectories to be extracted and forecasted at the same time from raw video inputs. The framework is able to extract 3D position and concurrently make accurate future predictions by modeling the complex socio-temporal interaction. The strengths of the model have been proven to be in congested highway driving scenarios and longer predictions, while the limitations lie in the position estimation and lane change scenarios. Further work can focus on implementing more efficient 3D position estimation techniques and possibly also implementing driving style into the prediction in order to improve the performance on lane change scenarios. Further suggestions may include focusing on more diverse driving scenarios such as urban driving with other road participants such as pedestrians and cyclists. 


\begin{thebibliography}{00}




















\bibitem{1} K. Messaoud, I. Yahiaoui, A. Verroust-Blondet, and F. Nashashibi, ``Attention based Vehicle Trajectory Prediction,'' \emph{IEEE Trans. Intell. Vehicles}, vol. 6, pp. 175--185, 2021.

\bibitem{20} S. Hochreiter and J. Schmidhuber, ``Long Short-Term Memory,'' \emph{Neural Comput.}, vol. 9, pp. 1735--1780, 1997.

\bibitem{21} J. Chung, C. Gulcehre, K. Cho, and Y. Bengio, ``Empirical Evaluation of Gated Recurrent Neural Networks on Sequence Modeling,'' \emph{arXiv}, arXiv:1412.3555, 2016.

\bibitem{22} M. Althoff and A. Mergel, ``Comparison of Markov Chain Abstraction and Monte Carlo Simulation for the Safety Assessment of Autonomous Cars,'' \emph{IEEE Trans. Intell. Transp. Syst.}, vol. 12, pp. 1237--1247, 2011.

\bibitem{44} Y. Xing, C. Lv, X. Mo, Z. Hu, C. Huang, P. Hang, and D. Cao, ``Toward Safe and Smart Mobility: Energy-Aware Deep Learning for Driving Behavior Analysis and Prediction of Connected Vehicles,'' \emph{IEEE Trans. Intell. Transp. Syst.}, vol. 22, pp. 4267--4280, 2021.

\bibitem{45} Y. Xing, C. Huang, C. Lv, L. Yahui, W. Hong, and D. Cao, ``A Personalized Deep Learning Approach for Trajectory Prediction of Connected Vehicles,'' \emph{SAE International Journal of Advances and Current Practices in Mobility}, 2020-01-0759, 2020.

\bibitem{35} C. Yu, X. Ma, J. Ren, H. Zhao, and S. Yi, ``Spatio-Temporal Graph Transformer Networks for Pedestrian Trajectory Prediction,'' \emph{Computer Vision--ECCV}, vol. 12357, pp. 507--523, 2020.

\bibitem{9} K. Chen, G. Chen, D. Xu, L. Zhang, Y. Huang, and A. Knoll, ``NAST: Non-autoregressive Spatial-Temporal Transformer for Time Series Forecasting,'' \emph{arXiv}, arXiv:2102.05624, 2021.

\bibitem{7} A. Vaswani, N. Shazeer, N. Parmar, J. Uszkoreit, A. N. Gomez, L. Kaiser, and I. Polosukhin, ``Attention Is All You Need,'' \emph{IEEE Adv. Neural Inf. Process. Syst.}, pp. 5998--6008, 2017.

\bibitem{10}
Next Generation Simulation (NGSIM). Available online: \url{http://ops.fhwa.dot.gov/trafficanalysistools/ngsim.html} (accessed on 20 September 2022).

\bibitem{16}
J. Xue, J. Fang, T. Li, B. Zhang, P. Zhang, Z. Ye, and J. Dou, ``Blvd: Building a Large-Scale 5D Semantics Benchmark for Autonomous Driving," \emph{arXiv}, vol. 1903.06405, 2019.

\bibitem{3}
N. Deo and M. M. Trivedi, ``Convolutional Social Pooling for Vehicle Trajectory Prediction,'' \emph{IEEE/CVF Conference on Computer Vision and Pattern Recognition Workshops (CVPRW)}, pp. 1468--1476, 2018.

\bibitem{92f}
N. Wojke, A. Bewley, and D. Paulus, ``Simple online and realtime tracking with a deep association metric,'' \emph{Proceedings of the 2017 IEEE International Conference on Image Processing (ICIP)}, pp. 3645--3649, 2017.

\bibitem{1p}
W. Luo, J. Xing, A. Milan, X. Zhang,W. Liu, T. Kim ``Multiple Object Tracking: A Literature Review,'' \emph{arXiv}, arXiv:1409.7618v5, 2022.

\bibitem{97f}
A. Mousavian, D. Anguelov, J. Flynn, and J. Kosecka, ``3D Bounding Box Estimation Using Deep Learning and Geometry,'' \emph{Proceedings of the IEEE Conference on Computer Vision and Pattern Recognition}, pp. 7074--7082, 2017.

\bibitem{mp}
F. Hasan, H. Huang, ``MALS-Net: A Multi-Head Attention-Based LSTM Sequence-to-Sequence Network for Socio-Temporal Interaction Modelling and Trajectory Prediction,'' \emph{Sensors}. vol. 23(1), pp. 530, 2023.

\bibitem{37}
S. Santurkar, D. Tsipras, A. Ilyas, and A. Madry, ``How Does Batch Normalization Help Optimization?,'' \emph{arXiv}, vol. 1805.11604, 2018.

\bibitem{39}
R. Pascanu, T. Mikolov, and Y. Bengio, ``On the difficulty of training recurrent neural networks,'' \emph{arXiv}, vol. 1211.5063, 2012.

\bibitem{40}
D. P. Kingma and J. Ba, ``Adam: A Method for Stochastic Optimization,'' \emph{arXiv}, vol. 1412.6980, 2014.

\bibitem{110f}
A. Dosovitskiy, G. Ros, F. Codevilla, A. Lopez, and V. Koltun, ``CARLA: An Open Urban Driving Si

\end{thebibliography}
\end{document}